\documentclass[conference]{IEEEtran}
\usepackage[T1]{fontenc}
\usepackage{balance}
\usepackage{cite,amsmath,amssymb,booktabs,graphicx,array,url}
\usepackage[hidelinks]{hyperref}
\usepackage{tikz}
\usetikzlibrary{arrows.meta}
\hypersetup{pdftitle={Persistent Identity in Person Tracking: Online Enrollment and Registered Recognition},pdfauthor={Ratha Siv, Matei Mancas, Bernard Gosselin}}

\title{Persistent Identity in Person Tracking: Online Enrollment and Registered Recognition}
\author{\IEEEauthorblockN{Ratha Siv, Matei Mancas, Bernard Gosselin}
\IEEEauthorblockA{University of Mons, Belgium}}
\IEEEoverridecommandlockouts
\IEEEaftertitletext{\vspace{-12pt}}

\begin{document}
\maketitle

\begin{abstract}
We extend PoseTReID with online appearance-bank matching (Gallery) and a residual-conflict guard for registered SVM recognition. All methods use supplied ground-truth (GT) person boxes. We average all-box assessment IDF1/HOTA equally across videos. On six SportsMOT clips, StrongSORT leads standalone trackers ($79.44/84.72\%$) and exceeds both Gallery variants. Over eleven videos, Gallery AIN ($83.29/85.44\%$) and COPE ($84.66/86.38\%$) exceed standalone trackers with complete results. This lead depends on two retrospectively added GTA development compilations and is absent on the original nine videos. Registered SVMs lead both cohorts among complete results, with AIN at $95.72/94.82\%$ over eleven videos. They use every GT crop from approximately $20\%$ of frames, including future appearances. On MOT20-03, registered AIN reached $96.80\%$ IDF1 and assigned the correct enrolled name to $90.72\%$ of assessment GT observations. Gallery and standalone trackers receive no per-video labelled enrollment. Both identity routes restore identities across internal title cards, but Gallery trails SORT on three videos (four in HOTA) and both SVMs trail SORT on MOT16-04/MOT20-03. Under six-clip persistent-only scoring, native Gallery gains $2.39$ percentage points in mean IDF1 over a matched control, with lower correct-P coverage and $5.02$ times as many encoder attempts. Pose gating is mixed; the guard's metric effect is not isolated.
\end{abstract}

\begin{IEEEkeywords}
person re-identification, persistent identity, multi-object tracking, online enrollment, identity conflicts, evaluation
\end{IEEEkeywords}

\section{Introduction}
Persistent identity links a person's encounters across local tracks. PoseTReID couples detection, tracking and re-identification (ReID)~\cite{posetreid2020}. Its extension study uses RBF-kernel SVMs for registered recognition and introduces the open-source \mbox{pyppbox} research toolbox~\cite{posetreid2022}.

We ask two questions. \emph{RQ1:} how do assignment and admission policies trade association quality against correct-P coverage and encoder work? \emph{RQ2:} do strong association scores imply correct registered names?

We contribute Gallery's online enrollment and optional pose-based admission, a guard for duplicate names after PoseTReID's rechecks (Fig.~\ref{fig:identitylayer}), and continuity, recognition, abstention and work measurements on eleven videos, including six SportsMOT policy clips~\cite{sportsmot2023}.

\section{Related Work and Scope}
SORT~\cite{sort2016} hosts our identity layer. DeepSORT~\cite{deepsort2017} adds appearance and a matching cascade to SORT's one-to-one assignment. ByteTrack~\cite{bytetrack2022} associates across confidence levels. BoT-SORT~\cite{botsort2022} combines motion, camera compensation and optional appearance. StrongSORT~\cite{strongsort2023} upgrades DeepSORT; we exclude offline AFLink/GSI.

PoseTReID queries new/unknown people and holders of duplicate IDs~\cite{posetreid2020,posetreid2022}; our guard follows those rechecks. Deep OC-SORT adapts appearance updates and matching weights to confidence and ambiguity~\cite{deepocsort2023}. Prior work studies quality-aware aggregation~\cite{qan2017} and selective extraction~\cite{selective2024}. MICRO-TRACK uses a circular-buffer gallery with centroids~\cite{microtrack2024}; HMP combines long-term prototypes and short-term queues~\cite{hmp2026}; Casao et al. select samples using uncertainty and diversity~\cite{casao2023}. Deferred commitment also limits early identity errors~\cite{buffered2026}. Gallery integrates exact-sample banks, temporary-to-persistent promotion, unique ownership and optional pose admission above a fixed host. We claim no individual ingredient as new and do not experimentally compare these memory designs.

Open-set ReID separates membership from identity selection~\cite{openset2014}; open-world evaluation motivates cross-time evidence~\cite{owd2024}. Consistent track IDs need not be correct names. Preprocessing, pretraining, enrollment and association confound architecture rankings.

\section{Persistent Identity Layer}
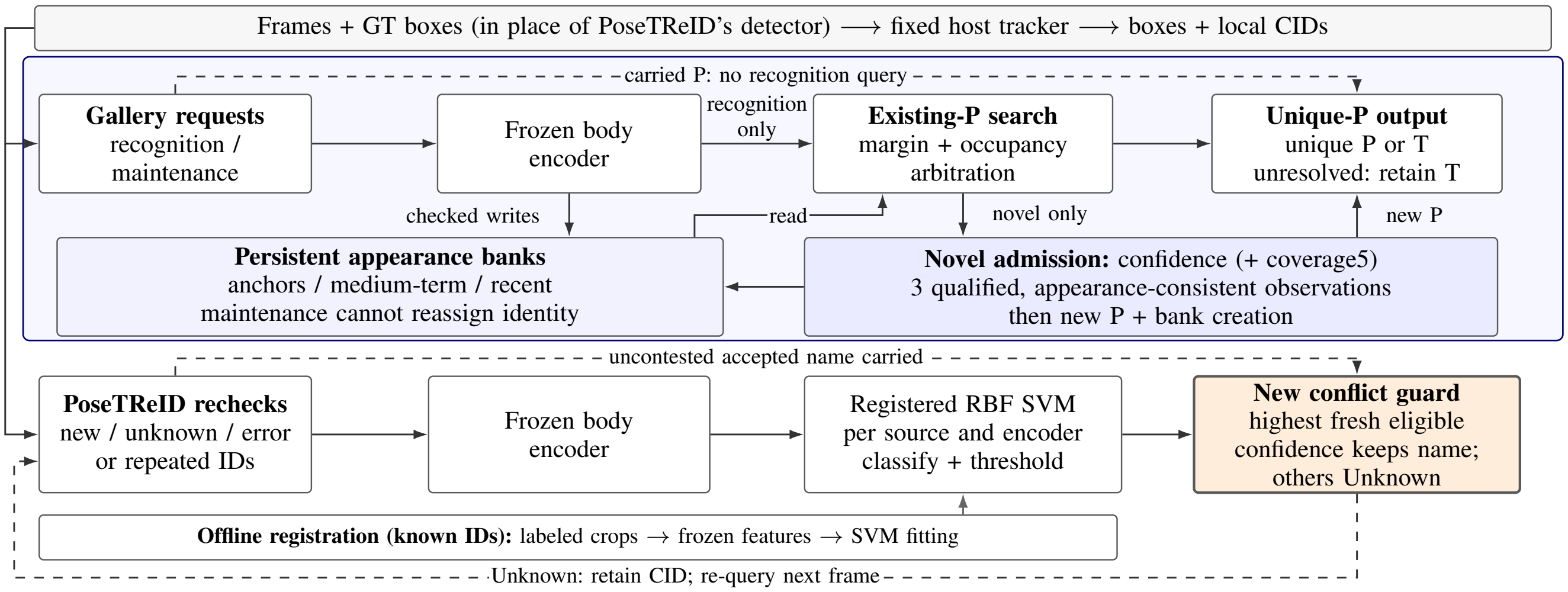
\begin{figure*}[t]
\centering
\begin{tikzpicture}[x=1mm,y=1mm,
  box/.style={draw=black!65,rounded corners=1.4pt,line width=.45pt,
    fill=white,align=center,inner xsep=2pt,inner ysep=2pt,font=\footnotesize},
  flow/.style={-{Latex[length=1.6mm,width=1.1mm]},line width=.55pt,draw=black!80},
  state/.style={flow,dashed},
  note/.style={font=\scriptsize,fill=white,inner sep=.6pt,align=center}]
  \node[box,text width=168mm,minimum height=5mm,fill=black!3] (host) at (89,57.4)
    {Frames + GT boxes (in place of PoseTReID's detector) $\longrightarrow$ fixed host tracker
     $\longrightarrow$ boxes + local CIDs};
  \draw[draw=blue!45!black,fill=blue!3,rounded corners=2pt,line width=.5pt]
    (3,22.5) rectangle (175,54.2);
  \node[box,text width=29mm,minimum height=11mm] (gsel) at (20,44.5)
    {\textbf{Gallery requests}\\recognition /\\maintenance};
  \node[box,text width=28mm,minimum height=11mm] (genc) at (64,44.5)
    {Frozen body\\encoder};
  \node[box,text width=32mm,minimum height=11mm] (gquery) at (108,44.5)
    {\textbf{Existing-P search}\\margin + occupancy\\arbitration};
  \node[box,text width=31mm,minimum height=11mm] (gout) at (152,44.5)
    {\textbf{Unique-P output}\\unique P or T\\unresolved: retain T};
  \node[box,text width=73mm,minimum height=11mm,fill=blue!5] (banks) at (44,28.5)
    {\textbf{Persistent appearance banks}\\anchors / medium-term / recent\\maintenance cannot reassign identity};
  \node[box,text width=76mm,minimum height=11mm,fill=blue!8] (admit) at (129,28.5)
    {\textbf{Novel admission:} confidence (+ coverage5)\\
     3 qualified, appearance-consistent observations\\then new P + bank creation};
  \draw[flow] (host.west) -- (1,57.4) -- (1,44.5) -- (gsel.west);
  \draw[flow] (gsel) -- (genc);
  \draw[flow] (genc) -- node[note,above,fill=none] {recognition\\only} (gquery);
  \draw[flow] (gquery) -- (gout);
  \draw[state] (gsel.north) -- (20,52)
    -- node[note,fill=blue!3] {carried P: no recognition query} (152,52)
    -- (gout.north);
  \draw[flow] (genc.south) -- (banks.north -| genc.south);
  \node[note,anchor=east,fill=none] at (61,36.5) {checked writes};
  \draw[flow] (banks.north -| 78,0) -- (78,36.5) -- (99,36.5)
    -- (gquery.south -| 99,0);
  \node[note,fill=blue!3] at (88.5,36.5) {read};
  \draw[flow] (gquery.south) -- (admit.north -| gquery.south);
  \node[note,anchor=west,fill=none] at (111,36.5) {novel only};
  \draw[flow] (admit.west) -- (banks.east);
  \draw[flow] (admit.north -| gout.south) -- (gout.south);
  \node[note,anchor=west,fill=none] at (155,36.5) {new P};
  \node[box,text width=29mm,minimum height=13mm,fill=white] (rsel) at (20,12)
    {\textbf{PoseTReID rechecks}\\new / unknown / error\\or repeated IDs};
  \node[box,text width=30mm,minimum height=13mm,fill=white] (renc) at (64,12)
    {Frozen body\\encoder};
  \node[box,text width=34mm,minimum height=13mm,fill=white] (rsvm) at (108,12)
    {Registered RBF SVM\\per source and encoder\\classify + threshold};
  \node[box,text width=35mm,minimum height=13mm,fill=orange!14,line width=.8pt] (rguard) at (152,12)
    {\textbf{New conflict guard}\\highest fresh eligible\\confidence keeps name;\\others Unknown};
  \draw[flow] (1,44.5) -- (1,12) -- (rsel.west);
  \draw[flow] (rsel) -- (renc);
  \draw[flow] (renc) -- (rsvm);
  \draw[flow] (rsvm) -- (rguard);
  \draw[state] (rsel.north) -- (20,20.5)
    -- node[note] {uncontested accepted name carried} (152,20.5)
    -- (rguard.north);
  \draw[state] (rguard.south) -- (152,-4)
    -- node[note] {Unknown: retain CID; re-query next frame} (2,-4)
    -- (2,9) -- (rsel.west |- 2,9);
  \node[box,text width=119mm,minimum height=5mm,fill=white,font=\scriptsize]
    (fit) at (65,.5)
    {\textbf{Offline registration (known IDs):} labeled crops $\rightarrow$ frozen features $\rightarrow$ SVM fitting};
  \draw[state,draw=black!60]
    (fit.north -| rsvm.south) -- (rsvm.south);
\end{tikzpicture}
\caption{Alternative identity routes in PoseTReID~\cite{posetreid2020,posetreid2022}. GT boxes replace detection; host associations are unchanged. The registered branch retains rechecks and offline SVM registration. Gallery (blue) adds online enrollment and optional coverage5 admission; the orange guard resolves residual name conflicts. Dashed arrows show propagation, retry and classifier supply. Standalone trackers are separate systems.}
\label{fig:identitylayer}
\end{figure*}
\subsection{Identity state, matching and memory}
Local continuity keys (CIDs) represent host associations. Gallery immediately assigns temporary labels T; persistent labels P index appearance banks. Each bank retains at most three anchor, three medium-term and three recent normalized body descriptors, $q=f_\theta(x)/\|f_\theta(x)\|_2$. Similarity is $S_i(q)=\max_{v\in V_i}q^\top v$. Winners must reach $\tau_m$, exceed the runner-up and satisfy margin $\gamma=0.05$. With one bank, the runner-up score is $-1$. Search uses exact samples; scaling is not evaluated.

Gallery searches existing banks first. Native and Coverage5 enroll after three consecutive valid observations on a stable unique key, with appearance consistency at $\tau_m$ and detector confidence $\geq0.75$. Reliable updates require similarity $\geq0.80$ and respect bank limits. Sparse maintenance cannot reassign P. Without periodic refresh, carried P can skip queries. Unresolved queries may request more features. Per-identity storage is bounded. Incorrect crops can contaminate memory.

Native \texttt{active\_no\_pose} reserves P for uniquely bound, continuous holders. A free bank goes to its unique top claimant. An occupied best match blocks fallback. A free rival at or above $\tau_m$ clears admission evidence. Otherwise a qualified newcomer can build its own P. New banks recruit only in later frames. Ties and unresolved conflicts retain T. Promotion preserves history. Unique labels can still be wrong or inherit tracker switches.

\subsection{Separate body-structure gate}
Coverage5 gates new enrollment only. Joints must fall inside the box and image with confidence $\geq0.5$. At least three shoulder/hip and two knee/ankle joints must span all four regions. Median heights must place shoulders above hips, and hips above knees and ankles. Head evidence is optional. This gate approximates body visibility.

Coverage5 requires a passing pose check on each admission observation. Failed structure checks retry after five frames; successes apply only to the current frame. Fresh YOLO26~\cite{ultralytics2026} nano-pose estimates use each region of interest at resolution $256$, without GT keypoints.

\subsection{Matched appearance-assignment control}
The matched control retains SORT, crops, frozen AIN, carried-P ownership, enrollment, bank limits, maintenance and Gallery's output-uniqueness check. Assignment and acceptance change jointly. Linear assignment minimizes summed $0.70-S_i(q)$ over permitted edges with $S_i(q)\geq0.70$, with one zero-cost unmatched option per query. This maximizes threshold surplus rather than the number of matches. Occupied banks remain reserved. New banks cannot recruit other newcomers in the same frame. Losing queries retain T. Stable CID and bank ordering with SciPy 1.17.1~\cite{scipy2020} resolves ties without jitter. Native and coverage5 require a $0.05$ rival margin, unlike the control. Shared thresholds therefore need not yield equal false-acceptance risk, histories or feature workloads.

\section{Protocol and Measurements}
\subsection{Frozen inputs and operating points}
Table~\ref{tab:cohort} lists six SportsMOT sources with clip-local IDs. Saved pilot results cover six different clips, two hosts, three encoders, three policies, 4,056 frames and 73 IDs (108 conditions), with return diagnostics. Its independent-matching control differs from the present matched control and registered SVM recognition.

\begin{table}[t]
\centering\footnotesize
\setlength{\tabcolsep}{3.5pt}
\caption{Six follow-up SportsMOT clips at 25 FPS. B/F/V denote basketball/football/volleyball. IDs are defined per clip.}
\label{tab:cohort}
\begin{tabular}{@{}llrrr@{}}
\toprule
Code & Sequence & Frames & GT rows & IDs \\
\midrule
B1 & \texttt{v\_4r8QL\_wglzQ\_c001} & 886 & 8183 & 10 \\
B2 & \texttt{v\_BgwzTUxJaeU\_c008} & 500 & 4475 & 10 \\
F1 & \texttt{v\_G-vNjfx1GGc\_c600} & 500 & 5462 & 19 \\
F2 & \texttt{v\_i2\_L4qquVg0\_c010} & 758 & 8262 & 21 \\
V1 & \texttt{v\_9MHDmAMxO5I\_c006} & 269 & 2601 & 12 \\
V2 & \texttt{v\_cC2mHWqMcjk\_c007} & 681 & 7877 & 12 \\
\bottomrule
\end{tabular}
\end{table}

SORT settings follow~\cite{posetreid2022}: maximum age $1$, minimum hits $3$ and intersection-over-union (IoU) $0.3$. The pilot's Centroid host uses spread $64$. Matching, update and margin thresholds are $0.70/0.80/0.05$, respectively. GT boxes carry confidence $1.0$. GT identity, team, row order, visibility and supplied representative points are removed before tracking. Box-derived points and chronological frames are retained. GT rows with ID zero remain in evaluation. Runs use fresh processes to preserve paired host geometry and causal state.

Supplied IDs and boxes define GT: ``correct'' means agreement with these annotations.
Crops and scoring retain annotation errors, occlusions, look-alike people and overlapping people inside boxes.

OSNet/AIN~\cite{osnet2019,osnetain2022} use frozen Torchreid~\cite{torchreid2019} \texttt{ms\_d\_c} weights pretrained on MSMT17, DukeMTMC-reID and CUHK03. Image-only COPE~\cite{cope2026} uses Occluded-Duke stride-16 weights. COPE's normalized 1280D descriptor combines 768D BN and 512D projected CLS features. Its crop preprocessing uses OpenCV linear resizing to $128\times256$, RGB/PIL bicubic resizing at the same size, and mean and standard deviation $0.5/0.5$. Policy runs use no fitted identity classifier, re-ranking or precomputed embeddings.

\subsection{Prospective six-source follow-up}
The SORT/AIN follow-up uses the native default encoder. Before new outputs, we froze sources, exclusions, sport-stratified hash order, models, implementation and endpoints. Six complete clips (two per sport) contain 3,594 frames, 36,860 GT observations and 84 clip-local IDs. Videos and visible matchups differ; distinct games/dates are unconfirmed. No source was replaced. Three policies yield 18 conditions with unchanged GT-box/host settings.

The primary comparison reports native-minus-control IDF1: equal-clip mean, clip differences and leave-one-clip-out means. Coverage5-minus-native is secondary; count-aware pools are separate. Any failed/invalid arm makes the summary incomplete. Source-sensitivity ranges are not population intervals.

A \emph{return} follows at least one second of GT absence; all 69 were listed
before scoring. Each policy's \emph{anchor} is the latest P associated with that GT identity
in the second before the gap. Two-second post-return diagnostics measure anchor
retention (T, Unknown and misses count as zero), reuse on another GT identity,
and consecutive duration. Missing anchors count as failures. Windows ending
at the next GT absence or clip end are marked.

\subsection{Eleven-video recognition and tracker comparison}
Six SportsMOT clips, bundled GTA5GP (GP, 958 frames), MOT16-04~\cite{mot2016}
(1,050), MOT20-03~\cite{mot202020} (2,405) and Normal/Hard all-in-one (AIO)
PoseTReID GTA compilations~\cite{posetreid2020} (GNA, 1,030; GHA, 3,741)
total 12,778 frames and 931 source-local IDs. GNA/GHA concatenate two/four
camera configurations from~\cite{posetreid2022} with 60-frame black-screen title
cards. This footage was used during development. GHA then GNA were added
retrospectively, freezing settings before each experiment.
Table~\ref{tab:comparison_allbox}'s methods share assessment masks with fresh no-ReID
SORT references. Paired streams verify unchanged CID/box geometry under both
identity routes; SORT settings and Gallery operating points remain unchanged.
Methods receive identical chronological images and anonymous confidence-one boxes.
MOT uses mark-1/class-1 targets without visibility filtering or official ignore
processing: these are not leaderboard scores. Confidence-one inputs do not
exercise ByteTrack's low-confidence recovery.

GNA/GHA's released 2023 annotations were semi-regenerated with YOLOv8, SORT,
OSNet-AIN and an unspecified custom component~\cite{posetreidannotations2023}.
GP uses YOLO26l/SORT with reviewed identity joins and temporal box corrections.
GNA/GHA share SORT and OSNet-AIN families with evaluated routes; GP shares SORT.
Provenance records are included.

For $N$ frames, $K=\lfloor N/5\rfloor$ nearly equal-length strata partition the timeline.
Each supplies the minimum SHA-256-ranked frame using a fixed seed, source ID,
stratum and frame indices. All methods/encoders share this deterministic
pseudorandom draw, not every fifth frame or per-identity draws.
Sampled GT IDs supply the SVM class labels.

Enrollment differs from~\cite{posetreid2022}: every GT box in our sampled registration
frames is cropped, without filtering or per-identity limits (93,180 crops;
Table~\ref{tab:full20_enrollment}).
Identities absent from these frames are not enrolled but are still scored.

A shared scoring mask excludes registration frames without resetting state,
plus one GNA frame (3 observations) and eight GHA frames (36) with duplicate GT IDs.
Tracking retains all boxes; training retains all 545 GNA and 2,894 GHA crops.
Enrollment includes future appearances and assessment-frame neighbors, so
these references cannot establish past-only or unseen-video generalization.
Gallery and external trackers receive no per-video labelled enrollment.
Each source/encoder pair uses a probability-enabled multiclass RBF SVM~\cite{scikitlearn2011}
($C=1$, \texttt{gamma=scale}), fitted then frozen. All 22 runs process complete sequences.
For native probabilities $p_k(x)$, names $y_k$ and
$k^*=\arg\max_k p_k(x)$, each valid fresh query proposes
\begin{equation}
\hat y(x)=\begin{cases}
y_{k^*}, & 100p_{k^*}(x)\geq\tau,\\
\mathrm{Unknown}, & \text{otherwise},
\end{cases}
\label{eq:svm_gate}
\end{equation}
with $\tau=35$ throughout, as for body ReID in~\cite{posetreid2022}. This gate
precedes label propagation and conflict handling.
Registered AIN uses raw features, whereas Gallery and COPE normalize features.

Registered inference enables \texttt{deduplicate} and \texttt{resolve\_conflicts}.
After selective and duplicate rechecks, the residual-conflict guard refreshes
stale contenders and resolves collisions by highest fresh confidence.
Ties favor the previous owner, then stable CID/order. Invalid/losing contenders
become Unknown, retain local handles and retry next frame.
This \emph{PoseTReID with a conflict guard} adds a step absent from the original.
Full-20\% enrollment and the guard are evaluated together; neither effect is isolated.

\subsection{Standalone implementations and operating points}
Standalone trackers use pinned author implementations. BoT-SORT's ablation
changes only its ReID flag. StrongSORT retains its appearance baseline,
enhanced-correlation-coefficient camera compensation~\cite{ecc2008}, noise-scale-adaptive
Kalman filter, exponential-moving-average appearance updates, motion cost and
matching without a cascade. Maximum age is 30 frames. Outputs require three-hit
confirmation and an update at most one frame old, allowing native one-frame predictions.
Camera compensation is causal, and offline AFLink/GSI are disabled.
The native measurement-noise scale remains zero for confidence-one inputs.

BoT-SORT and StrongSORT share the FastReID~\cite{fastreid2020} MOT17 SBS50 checkpoint and preprocessing:
clipped RGB crops at $128\times384$, in-model normalization and normalized
embeddings. StrongSORT is a shared-encoder adaptation, not an exact paper-weight
reproduction. MOT17 reuses MOT16 footage, including sequence 04~\cite{motchallenge2021}.
Its use for checkpoint training crops is unknown. Possible training overlap
limits transfer claims but cannot explain appearance-free results.
Settings are frozen before runs; hashes bind reused streams to recorded inputs.

\subsection{Three distinct identity measurements}
\emph{Persistent-only} policy scoring counts P as predictions and T/Unknown as abstentions. Same-frame duplicate P invalidate the contract, even on masked frames. Unrepaired N/A rows are retained; raw CIDs measure host continuity.

Eleven-video \emph{all-box} scoring retains every emitted box: Gallery P/T, registered R, Unknown/Error as stable local CIDs in namespace U, and external native IDs. No repair/interpolation is applied. TrackEval~\cite{trackeval2020} computes permutation-invariant IDF1~\cite{identity2016} and HOTA/AssA~\cite{hota2021}, which can credit Unknown-CID continuity without recognition. The contracts cannot be pooled or substituted.

\emph{Literal-name correct coverage} divides exact name matches by all assessment GT observations (correct, wrong, Unknown or missed). Unknown-person false acceptance uses all unenrolled-person GT observations. Correct-P scoring pairs boxes one-to-one per frame at IoU $\geq0.5$. Each run maps P labels one-to-one to GT identities to maximize agreement over the whole policy clip or comparison assessment frames. Maps never enter inference. Diagnostics retain splits, merges, never-recognized identities and the historical interface score.

\subsection{Computational cost}
Work counts encoder attempts and extra pose forwards after preparation and warmup. Policy runs use RTX 4090/PyTorch 2.5.1+cu121, FP32 with TF32 disabled. Registered encoders use GPU batches of up to 16 enrollment crops or one online query crop. SVM fitting and prediction use CPU. SBS50 uses FP16 batches of eight. Uncontrolled, unrepeated timings cannot rank speed.

\section{Results and Counter-Evidence}
\begin{figure*}[t]
\centering
\includegraphics[width=\textwidth]{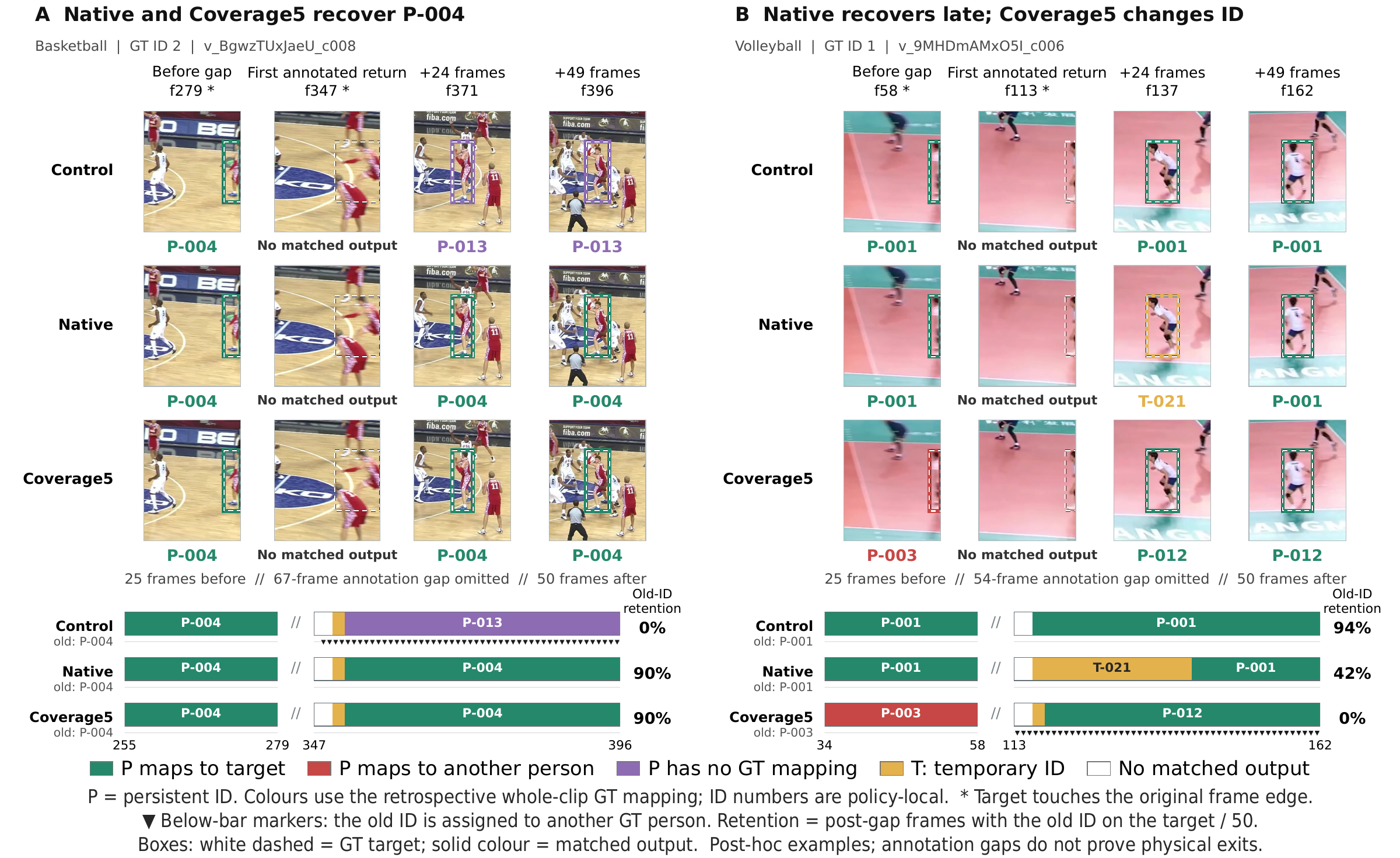}
\caption{Post-hoc SportsMOT examples~\cite{sportsmot2023}. In A, Control's first post-gap triangle appears at frame~348, when another GT person carries the old ID, one frame after the target's first annotated return.}
\label{fig:qualitative_sequence}
\end{figure*}
\subsection{Six-source policy trade-offs}
\begin{table}[t]
\centering\footnotesize
\setlength{\tabcolsep}{3pt}
\caption{Persistent-only IDF1 (\%) for all six follow-up clips. C=matched control, N=native, G=coverage5; differences (pp) use unrounded scores. Throughout, bold marks metric maxima or error/work minima before rounding, including ties.}
\label{tab:r8_clip_idf1}
\begin{tabular}{@{}lrrrrr@{}}
\toprule
Clip & C & N & G & $\Delta$N--C & $\Delta$G--N \\
\midrule
B1 & 81.35 & 81.38 & \textbf{83.44} & +0.04 & +2.06 \\
B2 & 86.77 & \textbf{93.36} & 92.80 & +6.59 & $-0.57$ \\
F1 & 52.30 & 54.90 & \textbf{57.56} & +2.60 & +2.66 \\
F2 & 51.74 & \textbf{56.89} & 54.93 & +5.14 & $-1.96$ \\
V1 & \textbf{77.04} & 76.70 & 76.20 & $-0.34$ & $-0.50$ \\
V2 & 74.65 & 74.98 & \textbf{77.23} & +0.33 & +2.25 \\
\bottomrule
\end{tabular}
\end{table}

\begin{table}[t]
\centering\footnotesize
\setlength{\tabcolsep}{3pt}
\caption{Six-clip count-aware pools (36,860 GT observations per policy), percentages. Raw CID is the common host reference; persistent-state columns do not apply. P is persistent-label coverage; C-P additionally requires mapped correctness. Policy metrics retain the persistent-only contract. Bold includes the host reference.}
\label{tab:confirmation_pooled_quality}
\begin{tabular}{@{}lrrrrr@{}}
\toprule
Policy & IDF1 & AssA & HOTA & P & C-P \\
\midrule
Raw CID & 68.82 & 59.29 & \textbf{75.27} & N/A & N/A \\
Control & 69.54 & 57.02 & 73.68 & \textbf{96.40} & \textbf{67.54} \\
Native & 72.75 & 64.03 & 73.76 & 86.08 & 66.88 \\
Coverage5 & \textbf{73.57} & \textbf{64.33} & 73.11 & 84.27 & 66.97 \\
\bottomrule
\end{tabular}
\end{table}

All 18 conditions completed with valid contracts, no same-frame duplicate P and no recorded extraction failure. Table~\ref{tab:r8_clip_idf1} reports IDF1 and paired differences. The equal-clip mean native-minus-control IDF1 difference is $+2.39$ pp. Differences are positive on five clips and $-0.34$ pp on one (range $[-0.34,+6.59]$). All six leave-one-clip-out means are positive ($[+1.55,+2.94]$). These are descriptive source-sensitivity results.

Count-aware pooling differs from equal-clip averaging (Table~\ref{tab:confirmation_pooled_quality}). Native improves IDF1 and AssA over the control, with nearly equal HOTA but lower P and correct-P coverage. Its HOTA remains below raw CID continuity.

\begin{table}[t]
\centering\footnotesize
\setlength{\tabcolsep}{3pt}
\caption{Six-clip totals. Wrong counts P observations mapped to another GT person. Frag sums distinct matched P labels beyond the first per GT person. Impure counts shared P labels and Never counts people without P, per clip. Body/Pose counts encoder/additional pose forwards. Single-pass timing includes logging.}
\label{tab:confirmation_identity_work}
\begin{tabular}{@{}lrrrrrr@{}}
\toprule
Policy & Wrong & Impure & Frag & Never & Body/Pose & ms/f \\
\midrule
Control & 8121 & 45 & 97 & \textbf{0} & \textbf{931}/\textbf{0} & \textbf{10.96} \\
Native & \textbf{3793} & \textbf{24} & 65 & 2 & 4669/\textbf{0} & 26.94 \\
Coverage5 & 4543 & 29 & \textbf{54} & 1 & 5356/546 & 31.07 \\
\bottomrule
\end{tabular}
\end{table}

\emph{RQ1:} native gains IDF1 with lower correct-P coverage and $4,669/931=5.02\times$ the control's encoder attempts, despite fewer impure labels and wrong-mapped observations (Table~\ref{tab:confirmation_identity_work}). Different policy states prevent isolation of assignment runtime or individual mechanisms.

Coverage5-minus-native equal-clip mean IDF1 is $+0.66$ pp. Differences are negative on three clips (range $[-1.96,+2.66]$), with leave-one-out means $[+0.26,+1.18]$. Pooled IDF1 rises, but HOTA falls and P coverage drops ($86.08\to84.27\%$). Correct-P coverage is nearly unchanged ($66.88\to66.97\%$). Wrong-mapped observations rise ($3,793\to4,543$), as do impure labels ($24\to29$), while fragments and never-P identities decrease. Encoder attempts rise ($4,669\to5,356$), with 546 additional pose forwards. These results do not establish a general benefit from pose gating.

Of the 69 returns, 56 have a full two-second window and 13 end sooner at the
next GT absence or clip end. Native reduces anchor reuse on
another GT identity ($10/69$ versus the control's $32/69$), but increases missing anchors
($15/69$ versus $2/69$) and lowers mean per-episode all-GT retention ($17.95\%$ versus $19.91\%$).
First reuse of the same P does not establish sustained correctness or long-absence recovery.

Figure~\ref{fig:qualitative_sequence} shows the first qualifying complete-window cases of improved native anchor retention and coverage5 anchor loss, selected in sequence/GT-ID/time order. Two-second own-anchor retention for control/native/coverage5 is 0/90/90\% (B2) and 94/42/0\% (V1). Selection is post-hoc, and saved evidence records all 69 eligibility decisions. In V1, coverage5 loses the old anchor but its new label agrees with GT, so anchor loss is not always an error.

\subsection{Named recognition with unfiltered enrollment}
\begin{table}[t]
\centering\footnotesize
\setlength{\tabcolsep}{3.4pt}
\caption{Unfiltered SVM enrollment. Frames: registration/scored. Enrolled/GT: enrolled/all local identities. Both SVMs train on all registration crops, including duplicates. Scoring excludes one additional GNA frame and eight GHA frames.}
\label{tab:full20_enrollment}
\begin{tabular}{@{}lrrr@{}}
\toprule
Source & Frames & Crops & Enrolled/GT \\
\midrule
B1 & 177/709 & 1,637 & 10/10 \\
B2 & 100/400 & 894 & 10/10 \\
F1 & 100/400 & 1,090 & 19/19 \\
F2 & 151/607 & 1,642 & 21/21 \\
V1 & 53/216 & 512 & 12/12 \\
V2 & 136/545 & 1,571 & 12/12 \\
M16 & 210/840 & 9,509 & 83/83 \\
M20 & 481/1,924 & 62,719 & 697/702 \\
GP & 191/767 & 10,167 & 54/54 \\
GNA & 206/823 & 545 & 3/3 \\
GHA & 748/2,985 & 2,894 & 5/5 \\
\midrule
Total & 2,553/10,216 & 93,180 & 926/931 \\
\bottomrule
\end{tabular}
\end{table}

\begin{table*}[t]
\centering\footnotesize
\setlength{\tabcolsep}{2.2pt}
\renewcommand{\arraystretch}{1.04}
\caption{GT-box assessment IDF1/HOTA (\%). GT counts scored observations. Registered SVMs use labelled crops from the $20\%$ frame sample, including future appearances. The $35\%$ threshold governs name acceptance. B/F/V: Table~\ref{tab:cohort}. GP: GTA5GP. M16/M20: MOT16-04/20-03. GNA/GHA: PoseTReID Normal/Hard AIO. Means weight complete videos equally (six SportsMOT or all eleven). StrongSORT ECC failed on GNA/GHA.}
\label{tab:comparison_allbox}
\begin{tabular}{@{}lr*{9}{c}@{}}
\toprule
Source & GT & SORT & \multicolumn{2}{c}{PoseTReID SVM (35\%)} & \multicolumn{2}{c}{Native Gallery} & ByteTrack & \multicolumn{2}{c}{BoT-SORT} & StrongSORT \\
 & & no ReID & AIN & COPE & AIN & COPE & & ReID off & ReID on & SBS50 \\
\midrule
B1 & 6,546 & 82.04/86.34 & 96.80/95.61 & \textbf{96.94}/\textbf{95.76} & 81.18/84.89 & 79.13/83.87 & 79.91/71.47 & 84.98/87.08 & 75.48/79.75 & 85.40/90.54 \\
B2 & 3,581 & 85.31/88.64 & 99.78/99.61 & \textbf{99.80}/\textbf{99.64} & 92.67/93.75 & 93.26/94.35 & 85.19/73.96 & 85.34/85.45 & 85.34/85.45 & 85.08/88.16 \\
F1 & 4,372 & 55.40/66.35 & 97.29/95.03 & \textbf{97.34}/\textbf{95.13} & 59.73/67.06 & 65.68/69.79 & 62.10/61.75 & 65.85/74.13 & 65.85/74.13 & 67.64/77.16 \\
F2 & 6,620 & 50.28/60.58 & \textbf{89.04}/\textbf{85.43} & 88.78/85.40 & 51.78/57.31 & 53.01/58.54 & 52.40/53.61 & 52.37/63.08 & 52.37/63.08 & 51.78/62.81 \\
V1 & 2,089 & 68.42/68.54 & 90.33/88.72 & 90.33/88.72 & 75.66/74.51 & 74.93/74.81 & 77.81/64.91 & 86.02/85.48 & 84.63/83.95 & \textbf{92.98}/\textbf{94.03} \\
V2 & 6,306 & 73.83/76.64 & 84.51/84.50 & 84.66/84.67 & 74.57/78.04 & 80.41/81.22 & 78.03/68.11 & 84.92/86.15 & 80.89/83.05 & \textbf{93.73}/\textbf{95.62} \\
M16 & 38,048 & 99.87/99.74 & 99.48/99.35 & 98.95/98.87 & 97.55/98.06 & 97.41/98.05 & 99.95/98.06 & \textbf{100.00}/\textbf{100.00} & \textbf{100.00}/\textbf{100.00} & 99.73/99.48 \\
M20 & 250,939 & \textbf{99.68}/99.40 & 96.80/96.92 & 94.16/94.77 & 85.76/89.93 & 89.49/92.64 & 99.00/94.19 & 99.60/\textbf{99.66} & 99.60/99.66 & 99.40/99.13 \\
GP & 40,850 & 99.38/99.52 & 99.93/\textbf{99.87} & 99.93/\textbf{99.87} & 99.79/99.73 & 99.79/99.74 & \textbf{99.96}/97.26 & 99.95/99.72 & 99.95/99.72 & 99.86/99.76 \\
GNA & 2,176 & 49.99/67.22 & \textbf{99.56}/\textbf{99.13} & \textbf{99.56}/\textbf{99.13} & 99.38/98.94 & 99.42/98.99 & 50.12/63.22 & 50.18/69.41 & 50.18/69.41 & N/A \\
GHA & 11,529 & 17.68/37.93 & \textbf{99.42}/\textbf{98.86} & \textbf{99.42}/\textbf{98.86} & 98.12/97.62 & 98.76/98.21 & 21.92/38.79 & 21.94/40.81 & 21.94/42.40 & N/A \\
\midrule
Mean (6) & --- & 69.21/74.51 & 92.96/91.49 & \textbf{92.97}/\textbf{91.55} & 72.60/75.93 & 74.40/77.10 & 72.57/65.64 & 76.58/80.23 & 74.09/78.24 & 79.44/84.72 \\
Mean (11) & --- & 71.08/77.35 & \textbf{95.72}/\textbf{94.82} & 95.44/94.62 & 83.29/85.44 & 84.66/86.38 & 73.31/71.39 & 75.56/81.00 & 74.20/80.06 & N/A \\
\bottomrule
\end{tabular}
\end{table*}

\begin{table}[t]
\centering\footnotesize
\setlength{\tabcolsep}{5pt}
\caption{Literal-name correct coverage (\%) with full-20\% enrollment and the fixed 35\% SVM gate. GT denotes all assessment observations: wrong names, Unknown and misses remain in the denominator. Source codes follow Tables~\ref{tab:cohort} and~\ref{tab:comparison_allbox}.}
\label{tab:r8_literal_names}
\begin{tabular}{@{}lrrr@{}}
\toprule
Source & GT & AIN & COPE \\
\midrule
B1 & 6,546 & 96.01 & \textbf{96.15} \\
B2 & 3,581 & 99.61 & \textbf{99.64} \\
F1 & 4,372 & 94.97 & \textbf{95.01} \\
F2 & 6,620 & \textbf{84.82} & 84.56 \\
V1 & 2,089 & \textbf{87.08} & \textbf{87.08} \\
V2 & 6,306 & 74.31 & \textbf{74.50} \\
M16 & 38,048 & \textbf{99.05} & 97.62 \\
M20 & 250,939 & \textbf{90.72} & 85.36 \\
GP & 40,850 & \textbf{99.86} & \textbf{99.86} \\
GNA & 2,176 & \textbf{99.13} & \textbf{99.13} \\
GHA & 11,529 & \textbf{98.86} & \textbf{98.86} \\
\bottomrule
\end{tabular}
\end{table}

On M20's 250,939 assessment observations (Table~\ref{tab:r8_literal_names}),
AIN yields 227,661 correct names, 878 wrong, 20,905 Unknown and 1,495 misses;
COPE yields 214,211 correct, 3,847 wrong, 31,386 Unknown and 1,495 misses.
Literal-name correct coverage is $90.72/85.36\%$ for AIN/COPE.
Full-stream emitted confidences, including carried values, span $0$--$70.83\%$ (AIN)
and $0$--$87.72\%$ (COPE). Five unenrolled IDs supply eight assessment
observations: both encoders give zero false names, one Unknown and seven misses,
too few to assess open-set performance.

\emph{RQ2:} strong association scores do not imply correct registered names.
Probability argmax differs from hard SVM prediction~\cite{libsvm2011}.
Rejected queries may rank the correct class first; accepted names may be wrong.
Effects of crop noise, class size, probability estimation and propagation are not separated.

M20's 62,719 training crops span 1--315 per class (median 85; six singletons).
Hard-prediction/probability-argmax accuracies are
$98.64/99.09\%$ (AIN) and $94.27/95.19\%$ (COPE); the $35\%$ gate accepts
47,885/37,514 crops. These resubstitution diagnostics are neither held-out nor causal tests.

\subsection{Same-host and external-tracker comparisons}
Table~\ref{tab:comparison_allbox} has 97 valid results. StrongSORT is N/A because
ECC fails after each GNA/GHA opening title card (frame 60). All 22 registered runs
have no same-frame duplicate names. All 44 identity-route runs preserve SORT CID/box geometry.

Both Gallery variants trail SORT in IDF1/HOTA on B1, M16 and M20, and in HOTA
on F2. M20 deficits reach $13.92/9.46$ pp for AIN and $10.20/6.76$ pp for COPE.

On GNA/GHA, both SVMs correctly name all 2,157/11,397 emitted assessment
boxes, with 19/132 misses. Post-hoc full-stream diagnostics, including enrollment
frames, show both routes restoring three/five prior identities after one/three
internal title cards despite new SORT CIDs. Completed standalone trackers assign new IDs.

BoT-SORT's two arms produce identical streams on six sources. ReID lowers both
metrics on B1, V1, V2 and M20 in assessment and preserved nine-source
full-stream scores, by less than $0.01$ pp on M20. On GHA, it raises HOTA by $1.6$ pp with unchanged IDF1.
Emission policies differ; full-stream and assessment scores overlap.
These comparisons cannot isolate appearance effects or rank general performance or speed.

\subsection{Discussion and limits}
On the original nine videos, both BoT-SORT arms and StrongSORT exceed both Gallery variants'
mean IDF1/HOTA; ByteTrack exceeds their IDF1, and SORT exceeds Gallery AIN's HOTA.
StrongSORT's eleven-video mean is N/A. SVMs use labelled future appearances.
The 60-frame cards exceed track-age thresholds (SORT: 1; ByteTrack and
BoT-SORT: 30). Audits confirm new standalone IDs at these transitions
but do not isolate causes of whole-video differences.

Historical 999-frame converted-selected MOT subsets favored native Gallery
over gating ($97.49$--$97.86\%$ versus $85.05$--$86.60\%$ persistent-only IDF1);
on historical GP, gating lowers this score from $99.82$--$99.89\%$ to $99.33\%$.
Evidence retains 33 duplicate-invalid pilot conditions and twelve N/A pools.

These MOT sequences have no same-ID annotation gaps. GNA/GHA title cards
do not establish physical returns. GT boxes, within-video
enrollment and shared annotation components limit end-to-end, past-only and
independent GTA validation.

Complete streams remain external; saved checks verify arithmetic, not every
score by replay. Code, results and data acquisition instructions are prepared
for release at \url{https://github.com/rathaROG/pyppbox-plus}.

\section{Conclusion}
With GT boxes and labelled enrollment from across each video, including future appearances, registered SVMs lead both cohort IDF1/HOTA means. Gallery needs no per-video labelled enrollment. Its eleven-video lead over standalone trackers with complete results depends on the two GTA development compilations.

Both routes restore identities across internal title cards despite new SORT CIDs. Native Gallery gains persistent-only IDF1 over the matched control with lower correct-P coverage and more encoder work. Pose gating gives mixed results. Future work includes guard ablation, past-only enrollment, controlled gaps and real detectors.

\section*{Acknowledgment}
The first author used OpenAI Codex and Anthropic Claude Code for source
inspection, tooling, evidence checks, drafting/revision (Sections I--VI),
figure code (Figs.~\ref{fig:identitylayer}--\ref{fig:qualitative_sequence}) and table formatting.
Authors retain responsibility; AI tools are not authors.

\clearpage
\balance
\bibliographystyle{IEEEtran}
\bibliography{references}
\end{document}